\documentclass[acmtog,nonacm]{acmart} 
\AtBeginDocument{%
  }

\renewcommand\footnotetextcopyrightpermission[1]{} %
\renewcommand\footnotetextauthorsaddresses[1]{} %

\usepackage{amsmath}
\usepackage{algorithm}
\usepackage{algorithmic}
\usepackage{graphicx}
\usepackage{enumitem}
\usepackage{amsmath}
\usepackage{algorithm}
\usepackage{algorithmic}
\usepackage{booktabs}
\usepackage{multirow}
\usepackage{multicol}
\usepackage[mathscr]{euscript}
\usepackage[dvipsnames]{xcolor}
\usepackage{colortbl}
\usepackage{makecell}
\usepackage{subcaption}
\usepackage[symbol]{footmisc}
\usepackage{wrapfig,lipsum,booktabs}
\usepackage{xcolor}   %
\usepackage{pifont}    %

\definecolor{ForestGreen}{RGB}{34,139,34}
\definecolor{myyellow}{RGB}{181, 181, 27}
\newcommand{\cmark}{\ding{51}}%
\newcommand{\xmark}{\ding{55}}%
\newcommand{\greencheck}{{\color{ForestGreen}\cmark}}

\newcommand{\redcheck}{{\color{red}\xmark}}
\usepackage[utf8]{inputenc} %
\usepackage[T1]{fontenc}    %
\usepackage{url}            %
\usepackage{booktabs}       %
\usepackage{amsfonts}       %
\usepackage{nicefrac}       %
\usepackage{microtype}      %
\usepackage[table]{xcolor}         %
\usepackage{multirow}
\usepackage{graphicx}

\newcommand\extrafootertext[1]{%
    \bgroup
    \renewcommand\thefootnote{\fnsymbol{footnote}}%
    \renewcommand\thempfootnote{\fnsymbol{mpfootnote}}%
    \footnotetext[0]{#1}%
    \egroup
}

\newcommand{\et}[2]{${#1}^{\pm{#2}}$}

\begin{document}
\fancyhf{} %
\pagestyle{empty} %

\definecolor{iccvblue}{rgb}{0.21,0.49,0.74}
\hypersetup{
  linkcolor=iccvblue,
  citecolor=iccvblue,
  urlcolor=iccvblue,
  filecolor=iccvblue
}

\title{UniMo: Unifying Human and Animal Motion Generation}

\author{Zeyu Zhang}
\authornote{Equal contribution.}
\affiliation{%
  \institution{The Australian National University}
  \country{Australia}}

\author{Zhiyuan Zhang}
\authornotemark[1]
\affiliation{%
  \institution{Adelaide University}
  \country{Australia}}

\author{Siheng Wang}
\authornotemark[1]
\affiliation{%
  \institution{Westlake University}
  \country{China}}

\author{Yiran Wang}
\affiliation{%
  \institution{The University of Sydney}
  \country{Australia}}

\author{Danning Li}
\affiliation{%
  \institution{The Hong Kong University of Science and Technology (Guangzhou)}
  \country{China}}

\author{Ian Reid}
\affiliation{%
  \institution{Mohamed bin Zayed University of Artificial Intelligence}
  \country{United Arab Emirates}}

\author{Richard Hartley}
\affiliation{%
  \institution{The Australian National University}
  \country{Australia}}

\begin{abstract}
  The conditional generation of 3D motion has emerged as a key research topic due to its wide applicability across robotics, AR/VR, gaming, and content creation. However, extending recent advances in text-driven human motion generation to the animal domain remains challenging due to two core limitations. First, animals exhibit highly diverse skeletal topologies, unlike the standard human structure, making unified modeling across species difficult and leading to inefficient per-species models. Second, existing animal motion datasets suffer from limited scale and annotation quality, constraining model performance. To address these challenges, we propose \textbf{UniMo}, a unified point cloud-based motion generation framework that bypasses topological discrepancies by converting parametric skeletons into unparametric representations, further enhanced by dynamic sampling that allocates more points to active joints. Additionally, we present \textbf{UniML3D}, a large-scale motion-language dataset spanning both human and animal categories, containing \textbf{145{,}907} motion sequences and \textbf{433{,}388} captions—over \textbf{102$\times$} larger than existing animal datasets. Our method achieves state-of-the-art results on UniML3D and three public benchmark including HumanML3D, KIT-ML, and AnimalML3D, demonstrating the feasibility and effectiveness of unified human-animal motion generation.
  Website:~\href{https://steve-zeyu-zhang.github.io/UniMo}{https://steve-zeyu-zhang.github.io/UniMo}.
\end{abstract}

\begin{teaserfigure}
  \includegraphics[width=\linewidth]{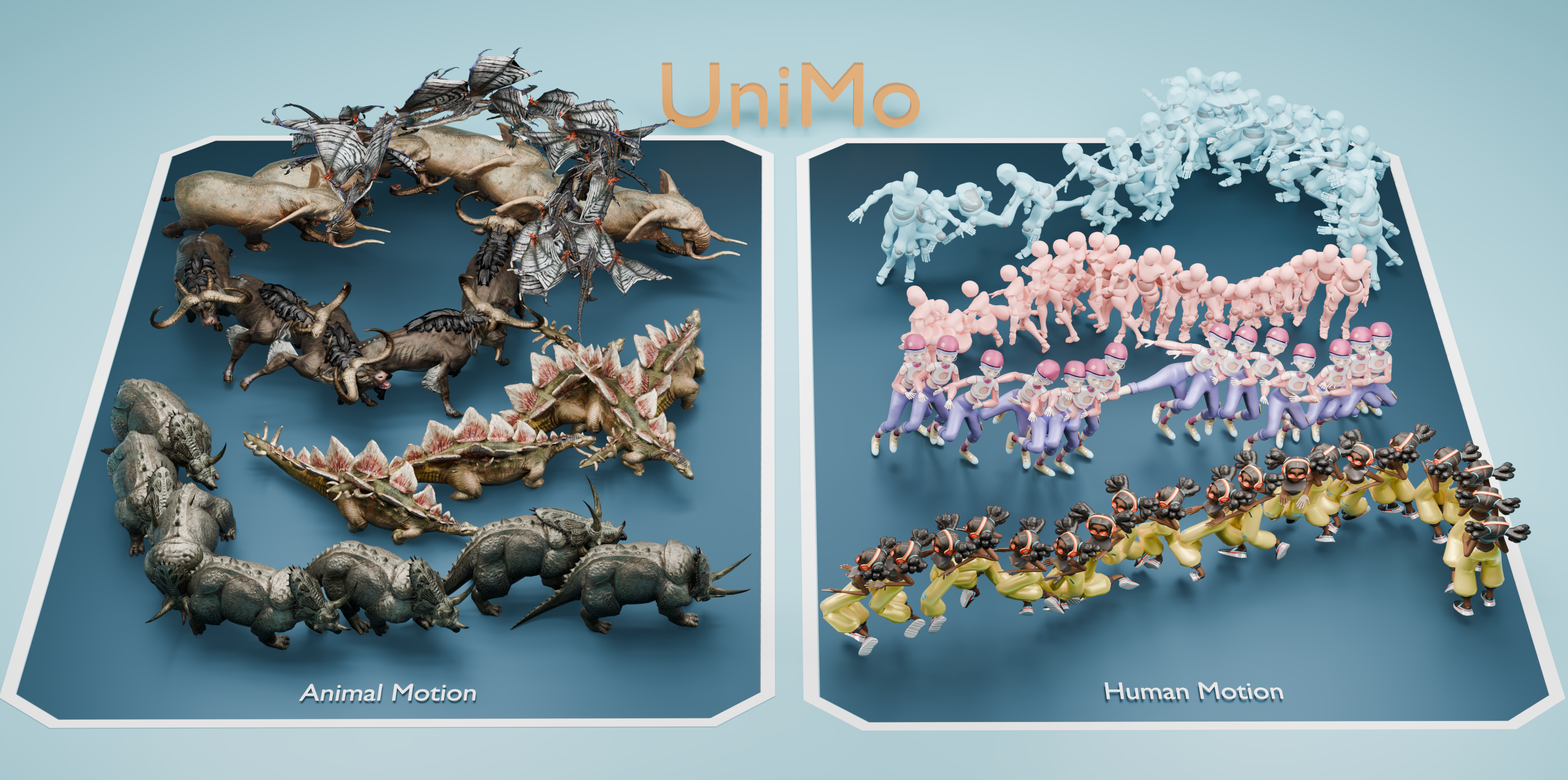}
  \caption{\textbf{UniMo} is a unified generative model capable of handling varying skeletal topologies, enabling unified human and animal motion generation within a single model.}
  \label{fig:teaser}
\end{teaserfigure}

\maketitle

\begin{figure*}[t]
    \centering
    \includegraphics[width=\linewidth]{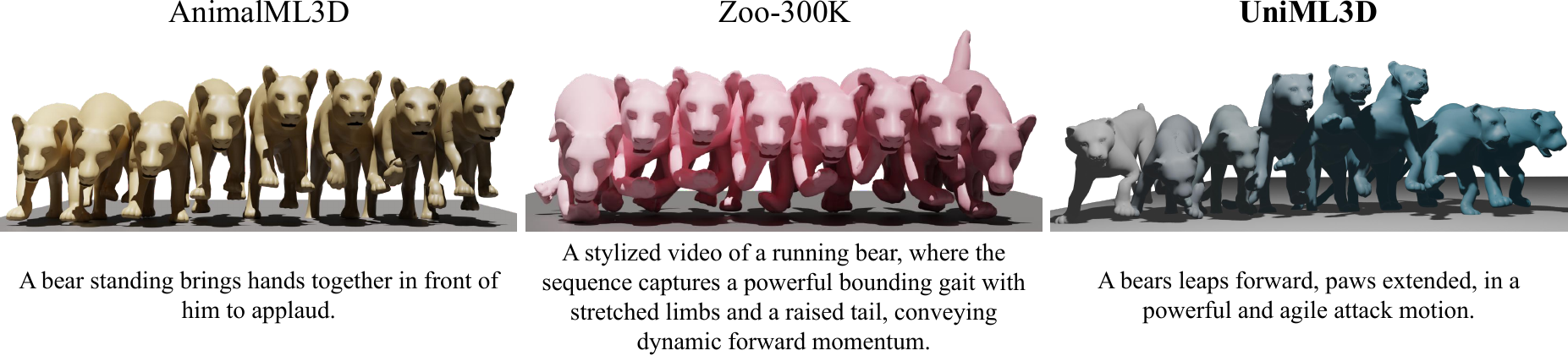}
    \caption{\textbf{Dataset comparsion.} AnimalML3D \cite{yang2024omnimotiongpt} contains unrealistic animal motions and limited captions, while Zoo-300K \cite{zhang2024motionavatar} suffers from low-quality motions and annotations with obvious hallucinations. In contrast, our UniML3D offers realistic motions and high-quality captions.}
    \label{fig:dataset}
\end{figure*}

\section{Introduction}
Recently, the conditional generation of 3D motions \cite{guo2020action2motion,petrovich2021action,guo2022generating,tevet2022motionclip} has attracted significant attention, as it is an important topic with a wide range of applications, including robot manipulation \cite{serifi2024robot}, urban planning \cite{chen2024omnire}, virtual \cite{ye2022neural3points} and augmented reality \cite{yue2021human}, game development \cite{de2020generating}, and video creation \cite{chen2024dreamcinema}. Especially, recent advances in text-driven human motion generation \cite{zhang2025motion,hong2024bipo,pinyoanuntapong2024mmm,hosseyni2024bad,pinyoanuntapong2024bamm,guo2024momask,yuan2024mogents} have led to breakthrough success in synthesizing realistic human motions from natural language descriptions, demonstrating the potential to greatly increase the accessibility and efficiency of motion animation. However, despite the success of human motion generation, two major challenges still hinder the adaptation of similar techniques to animal motion generation.

(1) \textbf{Varying topologies.} Unlike humans, whose topology can be represented with a standard parametric model \cite{xu2020ghum,alldieck2021imghum,loper2023smpl}, animals have a vast number of diverse species with highly varied morphologies, making it challenging to develop a unified generative model capable of handling a wide variety of skeletal topologies \cite{zuffi20173d}. This leads existing methods to train a separate model for each species, resulting in inefficiencies in both training, inference and storage~\cite{zhang2024motionavatar}.

(2) \textbf{Limited data.} Text driven animal motion generation is much less studied than human motion generation, primarily due to the lack of high quality datasets \cite{yang2024omnimotiongpt}. Animal motion data with text annotations, especially high quality and large scale datasets, are not available at a comparable scale to human motion datasets. Moreover, the poor quality of existing data limits the performance and potential of developing effective text to animal motion generation models, as shown in Figure \ref{fig:dataset}.

To address the first challenges, we present \textbf{UniMo}, a method that tackles diverse skeletal topology issues by converting parametric skeletons into an unparametric point cloud representation. This enables unified modeling of both human and animal motion within a single framework. We further introduce a dynamic point cloud sampling strategy that allocates more points to highly active joints based on motion dynamics.

To address the second challenge, we introduce \textbf{UniML3D}, a unified motion language dataset containing high quality motion sequences and carefully curated textual annotations for both human and animal categories, resulting in \textbf{145,907} motion sequences and \textbf{433,388} captions in total. Compared to the previous AnimalML3D dataset \cite{yang2024omnimotiongpt}, which contains only 1,240 motions and 3,720 captions, UniML3D is over \textbf{102$\times$} larger in both animal motion and caption scale.

To validate our effectiveness, we further conduct comprehensive experiments on both our UniML3D dataset and public benchmarks including HumanML3D, KIT-ML, and AnimalML3D. Our method outperforms previous state-of-the-art approaches, showing promising results for the future of unified human and animal motion generation.

The contributions of our paper can be summarized as follows:

\begin{itemize}[leftmargin=2em, itemsep=-0.1em, topsep=0em]
\item We present \textbf{UniMo}, a unified point cloud-based motion generation framework that handles diverse skeletal topologies across humans and animals via a dynamic sampling strategy. 
\item We construct \textbf{UniML3D}, the largest unified motion-language dataset to date, with 145,907 motions and 433,388 captions, over \textbf{102$\times$} larger than AnimalML3D.
\item Our method achieves state-of-the-art results on UniML3D and other three widely-used motion-language benchmarks: HumanML3D, KIT-ML, and AnimalML3D, consistently surpassing prior approaches in both text-motion alignment and motion quality across human and animal categories.
\end{itemize}

\section{Related Work}

\subsubsection{Animal motion generation.}

Previous methods for animal motion generation, such as OmniMotionGPT \cite{yang2024omnimotiongpt}, require a two-stage process during both training and inference, which first generating human motion, then converting it into animal motion via an implicit semantic mapping. This results in an inefficient pipeline and unrealistic animal motions that act like humans. Other animal motion generation \cite{zhang2024motionavatar} and editing \cite{li2023example,raab2023single,wang2025motiondreamer} methods require training a separate model for each category. Recent cross-category motion transfer \cite{zhang2024magicpose4d} and editing \cite{gat2025anytop} methods operate either through semantic alignment across joint groupings or by identifying joint pairs based on topological distances. However, they rely on reference motions and cannot generate arbitrary motions from text conditioning. T2M4LVO \cite{lee2025move} attempts to generate cross-category motion by flattening the token sequence over joints and frames with linear projection and position encoding, but this loses the spatio-temporal structure of motion. Additionally, the paper does not compare with the above baselines and is not yet open-sourced.

\begin{table*}[t]
\centering
\caption{\textbf{Datasets comparison.} Our UniML3D offers sufficient category and quantity, with more realistic motion and precise annotations.}
\label{tab:dataset}
\resizebox{0.8\linewidth}{!}{
\begin{tabular}{ccccccccc}
\toprule
Dataset & Human & Animal & Categories & Motions & Captions & R-Precision Top-3 $\uparrow$ & FID $\downarrow$ & MM-Dist $\downarrow$ \\ 
\midrule
HumanML3D \cite{guo2022generating} & \greencheck & \redcheck & 1 & 14,616 & 44,970 & 0.797 & 0.002 & 2.974\\
KIT-ML \cite{plappert2016kit} & \greencheck & \redcheck & 1 & 3,911 & 6,278 & 0.779 & 0.031 & 2.788\\
AnimalML3D \cite{yang2024omnimotiongpt} & \redcheck & \greencheck & 36 & 1,240 & 3,720 & 0.839 & 0.105 & 0.357\\
Truebones Zoo \cite{free} & \redcheck & \greencheck & 74 & 1,153 & - & - & 0.008 & - \\
Zoo-300K \cite{zhang2024motionavatar} & \redcheck & \greencheck & 65 & 270,254 & 270,254 & 0.767 & 0.024 & 2.559\\ 
\midrule
UniML3D (Animal) &   \redcheck  &    \greencheck & 31 & 127,380 & 382,140 & 0.841 & 0.013 & 2.193 \\ 
\textbf{UniML3D (Ours)} &   \greencheck  &    \greencheck & 32 & 145,907 & 433,388 & 0.846 & 0.011 & 2.458 \\ 
\bottomrule
\end{tabular}}
\end{table*}

\subsubsection{Human motion generation.}

Recent advances in human motion generation have effectively integrated diffusion \cite{zhang2024motion,zhang2024motiondiffuse,zhang2023remodiffuse} and VQ-VAE \cite{guo2024momask} based models to enable more realistic, diverse, and controllable motion synthesis. 
Foundational work like MDM \cite{tevet2022human} introduced a transformer-based diffusion approach for lifelike, text-driven motion generation. Expanding on this, MotionDiffuse \cite{zhang2024motiondiffuse} added refined control and diversity mechanisms, while MLD \cite{chen2023executing} boosted efficiency by operating within a latent space, reducing computational demands without sacrificing quality. Motion Mamba \cite{zhang2025motion} addressed the challenge of generating longer sequences, and ReMoDiffuse \cite{zhang2023remodiffuse} further enriched motion variability by incorporating retrieval-augmented diffusion. Meanwhile, autoregressive models like MoMask \cite{guo2024momask} enhanced temporal coherence through generative masked modeling, selectively revealing segments of the motion sequence. BAMM \cite{pinyoanuntapong2024bamm} introduced a bidirectional model to capture detailed motion with forward and backward dependencies. InfiniMotion \cite{zhang2024infinimotion} optimized transformer memory to support extended sequences, and KMM \cite{zhang2024kmm} prioritized essential frames to balance continuity and computational efficiency. MoGenTS \cite{yuan2024mogents} added spatial-temporal joint modeling for further structural consistency in generated motions.

\begin{figure*}[t]
    \centering    
    \includegraphics[width=\linewidth]{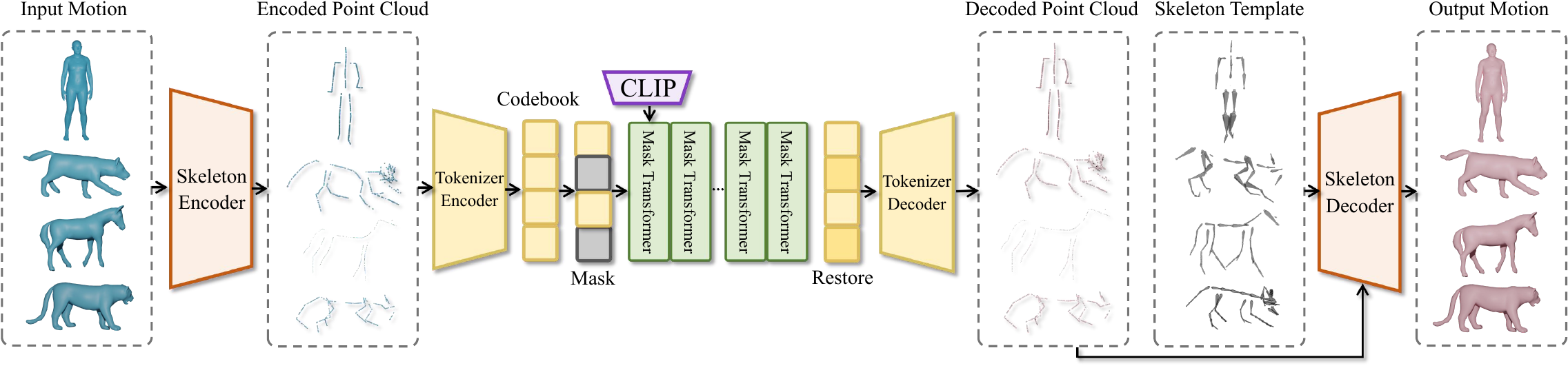}
    \caption{\textbf{UniMo architecture.} Given an input motion sequence, the skeleton encoder converts it into a point cloud representation, which is then discretized via a tokenizer and compressed using a codebook. The Mask Transformer performs masked modeling over the tokenized point cloud. The decoder reconstructs the motion through a two-stage process: decoding the point cloud and regressing back to skeletal motion. A frozen CLIP text encoder is used for text condition, and a skeleton template is applied to restore the final output motion.}
    \label{fig:arch}
\end{figure*}

\section{Datasets: UniML3D}

Currently, existing animal motion-language datasets suffer from either poor motion quality or inadequate text annotations, as shown in Figure~\ref{fig:dataset}. To address the challenge of lack of high-quality, unified human and animal motion-language datasets, we present the 3D Unified Motion-Language Dataset (UniML3D). For the UniML3D data synthetic pipeline, we select high-quality motion from Zoo-300K \cite{zhang2024motionavatar} with both quantitative metrics including R-Precision, FID, and MM-Dist, combining human expert evaluation to obtain motion sequences with 31 categories of animals. We then re-caption the selected high-quality motions using Qwen2.5-VL-72B-Instruct \cite{bai2025qwen2}, generating 10 captions per motion. Human experts select the top 3 captions for each motion as the final version. Each motion's caption selection is reviewed by at least two human experts to ensure inter-rater reliability. This results in a total of 127,380 animal motion sequences and 382,140 corresponding captions. We then combine them with human motions and captions from HumanML3D \cite{guo2022generating} and KIT-ML \cite{plappert2016kit}, resulting in 145,907 motion sequences and 433,388 captions in total. As shown in Table~\ref{tab:dataset}, our UniML3D provides a sufficient category and quantity of motion, and achieves significantly better quantitative metrics such as FID and R-Precision compared to previous animal motion-language datasets, indicating that our dataset contains more realistic motion and more precise text annotations.

\section{Methodology}

\subsection{Overview}
Our proposed framework consists of three key components: a \textit{Skeleton Encoder-Decoder}, a \textit{Point Cloud Tokenizer}, and a \textit{Mask Transformer} for sequence modeling. Given a motion sequence represented by joint positions and orientations, we first convert the structured skeleton into a point cloud via dynamic sampling based on joint motion magnitude. This allows the Skeleton Encoder to encode motions with varying topologies into a unified point-based representation. The point cloud is then compressed temporally using a VQ-VAE, where the Tokenizer Encoder maps each frame's point cloud to a latent vector and discretizes it through vector quantization. A Transformer-based Tokenizer Decoder reconstructs the original point cloud from the quantized sequence. To model temporal dependencies, we apply a Mask Transformer that learns to predict masked tokens in the quantized sequence using contextual information, enabling high-quality and diverse motion generation. Finally, the Skeleton Decoder reconstructs the full motion sequence from the predicted point cloud, guided by a template skeleton. Our framework enables unified human-animal motion generation with strong generalization across topological variations.

\begin{algorithm}[t]
\caption{Skeleton Encoder and Decoder}
\begin{algorithmic}[1]
\REQUIRE Motion sequence $(\mathbf{P},\mathbf{Q}) \in \mathbb{R}^{F\times J\times(3+4)}$, number of points $N$
\STATE Compute joint motion magnitude $m_{f,j}=\|\mathbf{p}_{f,j}-\mathbf{p}_{f-1,j}\|$
\STATE Dynamic sampling distribution 
$\pi_{f,j}=\frac{\|\mathbf{t}_j\|(1+\lambda m_{f,j})}{\sum_k\|\mathbf{t}_k\|(1+\lambda m_{f,k})}$
\STATE Sample joints $j\sim\text{Categorical}(\pi_f)$ and generate local points
$\mathbf{p}_{local}=\alpha\mathbf{t}_j+\boldsymbol{\epsilon}$
\STATE Transform to global coordinates
$\mathbf{p}_f=\mathbf{R}(\mathbf{q}_{f,j})\mathbf{p}_{local}+\mathbf{p}_{f,j}$
\STATE Form point cloud sequence $\mathbf{X}\in\mathbb{R}^{F\times N\times3}$
\STATE Predict keypoints $\mathbf{K}$ from $\mathbf{X}$
\STATE Solve shortest-arc quaternion IK to estimate rotations $\mathbf{Q}$
\STATE Recover joint positions via Forward Kinematics
\RETURN reconstructed motion $\mathbf{M}=(\mathbf{P},\mathbf{Q})$
\end{algorithmic}
\end{algorithm}
\subsection{Skeleton Encoder and Decoder}

Let a motion sequence $\mathbf{M} \in \mathbb{R}^{F \times J \times 7}$ be represented by joint positions $\mathbf{P} \in \mathbb{R}^{F \times J \times 3}$ and joint orientations $\mathbf{Q} \in \mathbb{R}^{F \times J \times 4}$, where $F$ denotes the number of frames, $J$ is the number of joints, and each joint is parameterized by a 3D coordinate for position and a unit quaternion for rotation.

\subsubsection{Skeleton Encoder}

To obtain a geometry-aware representation, we encode the articulated skeleton into a dense point cloud. Each joint $j$ is associated with a rest bone direction $\mathbf{t}_j \in \mathbb{R}^{3}$ that defines the direction of the bone extending from the joint. Points are sampled along these bones to form a spatial distribution around the articulated structure.

\textbf{\textit{Dynamic point cloud sampling.}} To adaptively allocate more points to dynamically moving joints, we introduce a motion-aware sampling distribution. For each frame $f$, we first estimate the motion magnitude of each joint as

\[
m_{f,j} = \|\mathbf{p}_{f,j} - \mathbf{p}_{f-1,j}\|.
\]

We then define a motion-aware categorical distribution that combines bone length and motion magnitude,

\[
j \sim \text{Categorical}(\pi_f), \qquad
\pi_{f,j} =
\frac{\|\mathbf{t}_j\|\,(1 + \lambda m_{f,j})}
{\sum_{k=1}^{J} \|\mathbf{t}_k\|\,(1 + \lambda m_{f,k})},
\]

where $\lambda$ controls the strength of motion-aware point allocation. This dynamic sampling assigns more point cloud samples to joints with higher motion frequency while maintaining a fixed total number of points.

Given the sampled joint, a point is generated along the corresponding bone segment by sampling a scalar $\alpha$ along the bone direction,

\[
\alpha \sim \mathcal{U}(0,1).
\]

The sampled point in the local joint coordinate system is

\[
\mathbf{p}_{\text{local}} = \alpha \mathbf{t}_j + \boldsymbol{\epsilon},
\]

where $\boldsymbol{\epsilon} \sim \mathcal{N}(\mathbf{0}, \sigma^2 \mathbf{I})$ is a Gaussian offset that provides spatial thickness around the skeleton.

To transform the point into the global coordinate system, we apply the joint rotation and translation. Let $\mathbf{R}(\mathbf{q}_{f,j})$ denote the rotation matrix corresponding to quaternion $\mathbf{q}_{f,j}$. The global point coordinate becomes

\[
\mathbf{p}_{f} = \mathbf{R}(\mathbf{q}_{f,j}) \mathbf{p}_{\text{local}} + \mathbf{p}_{f,j},
\]

where $\mathbf{p}_{f,j}$ is the global position of joint $j$ at frame $f$.

Repeating this process $N$ times produces a point cloud

\[
\mathbf{X}_f = \{\mathbf{p}_f^{(1)}, \mathbf{p}_f^{(2)}, \dots, \mathbf{p}_f^{(N)}\}, 
\qquad \mathbf{X}_f \in \mathbb{R}^{N \times 3},
\]

which provides a dense geometric encoding of the articulated skeleton at frame $f$. Stacking all frames yields the encoded motion representation

\[
\mathbf{X} \in \mathbb{R}^{F \times N \times 3}.
\]

This encoding converts the articulated structure into a spatial point distribution while preserving bone orientation and hierarchical motion structure through quaternion-based transformations.

\[
\mathbf{p}_f
=
\mathbf{R}(\mathbf{q}_{f,j}) \mathbf{p}_{\text{local}}
+
\mathbf{p}_{f,j}.
\]

\subsubsection{Skeleton Decoder}

From the encoded point cloud $\mathbf{X}$, a keypoint predictor estimates joint-associated keypoints 
$\mathbf{K} \in \mathbb{R}^{F \times J \times 3}$.
Given the encoded point cloud representation $\mathbf{X}$ and predicted keypoints $\mathbf{K}$, the goal of the decoder is to recover the articulated skeleton representation $\mathbf{M}$ by estimating joint rotations that align the rest skeleton with the observed geometric directions. We formulate this process as an \textit{inverse kinematics (IK)} problem solved using a shortest-arc quaternion formulation, followed by \textit{forward kinematics (FK)} to reconstruct joint positions.

\textbf{\textit{Inverse kinematics.}} For each frame $f$ and joint $j$, we first compute the target bone direction from the joint position to the decoded keypoint,

\[
\mathbf{v}_{f,j} = \mathbf{k}_{f,j} - \mathbf{p}_{f,j},
\]

where $\mathbf{p}_{f,j}$ denotes the global position of joint $j$ and $\mathbf{k}_{f,j}$ denotes the decoded keypoint location associated with the joint.

Let $\mathbf{t}_j$ denote the rest bone direction of joint $j$. The objective is to compute a quaternion rotation $\mathbf{q}_{f,j}$ such that

\[
\mathbf{R}(\mathbf{q}_{f,j}) \mathbf{t}_j \approx \mathbf{v}_{f,j}.
\]

To solve this alignment, both vectors are normalized,

\[
\mathbf{u}_j = \frac{\mathbf{t}_j}{\|\mathbf{t}_j\|}, 
\qquad
\mathbf{v}_{f,j} = \frac{\mathbf{v}_{f,j}}{\|\mathbf{v}_{f,j}\|}.
\]

We then compute the shortest-arc quaternion that rotates $\mathbf{u}_j$ to $\mathbf{v}_{f,j}$,

\[
\mathbf{q}^{\text{dir}}_{f,j}
=
\text{normalize}
\left(
\begin{bmatrix}
1 + \mathbf{u}_j^\top \mathbf{v}_{f,j} \\
\mathbf{u}_j \times \mathbf{v}_{f,j}
\end{bmatrix}
\right).
\]

This quaternion represents the minimal rotation that aligns the rest bone direction with the target direction. Since the direction constraint does not uniquely determine the rotation around the bone axis, we introduce an additional roll rotation around $\mathbf{u}_j$. Let $\theta_{f,j}$ denote the roll angle around the bone axis. The roll quaternion is defined as

\[
\mathbf{q}^{\text{roll}}_{f,j}
=
\begin{bmatrix}
\cos \frac{\theta_{f,j}}{2} \\
\sin \frac{\theta_{f,j}}{2}\,\mathbf{u}_j
\end{bmatrix}.
\]

The final joint rotation is obtained by composing the direction alignment and roll rotations,

\[
\mathbf{q}_{f,j}
=
\mathbf{q}^{\text{dir}}_{f,j}
\otimes
\mathbf{q}^{\text{roll}}_{f,j}.
\]

\textbf{\textit{Forward kinematics.}} After estimating joint rotations $\mathbf{Q}$, the global joint positions are reconstructed using \textit{forward kinematics (FK)} along the skeletal hierarchy. For each joint $j$ with parent $p(j)$, the joint position is computed recursively as

\[
\mathbf{p}_{f,j}
=
\mathbf{p}_{f,p(j)}
+
\mathbf{R}(\mathbf{q}_{f,p(j)}) \mathbf{o}_j,
\]

where $\mathbf{o}_j$ denotes the rest offset between joint $j$ and its parent.

Applying FK along the entire skeleton hierarchy yields the reconstructed articulated motion representation

\[
\mathbf{M} = (\mathbf{P}, \mathbf{Q}).
\]

\subsubsection{Learning Objective}

The Skeleton Encoder-Decoder is trained to reconstruct the articulated motion sequence from the encoded point cloud representation. The learning objective encourages geometric consistency between the reconstructed skeleton and the underlying point cloud while enforcing physically valid joint rotations.

\textbf{\textit{Keypoint reconstruction.}}
The keypoint predictor estimates joint-associated keypoints 
$\mathbf{K} \in \mathbb{R}^{F \times J \times 3}$ from the encoded point cloud $\mathbf{X}$. 
We supervise the predicted keypoints using the ground-truth joint positions:

\[
\mathcal{L}_{\text{kp}}
=
\frac{1}{FJ}
\sum_{f=1}^{F}\sum_{j=1}^{J}
\left\|
\mathbf{k}_{f,j} - \mathbf{p}_{f,j}
\right\|_2^2 .
\]

\textbf{\textit{Joint position reconstruction.}}
After estimating joint rotations through inverse kinematics and reconstructing joint positions via forward kinematics, we enforce consistency between the reconstructed joints and the ground-truth skeleton:

\[
\mathcal{L}_{\text{pos}}
=
\frac{1}{FJ}
\sum_{f=1}^{F}\sum_{j=1}^{J}
\left\|
\hat{\mathbf{p}}_{f,j} - \mathbf{p}_{f,j}
\right\|_2^2 ,
\]

where $\hat{\mathbf{p}}_{f,j}$ denotes the reconstructed joint position obtained after FK.

\textbf{\textit{Quaternion normalization.}}
To ensure valid rotations, we regularize the predicted quaternions to maintain unit norm:

\[
\mathcal{L}_{\text{quat}}
=
\frac{1}{FJ}
\sum_{f=1}^{F}\sum_{j=1}^{J}
\left(
\|
\mathbf{q}_{f,j}
\|_2 - 1
\right)^2 .
\]

\textbf{\textit{Overall objective.}}
The final training objective is defined as a weighted combination of these losses:

\[
\mathcal{L}_{\text{skel}}
=
\lambda_{\text{kp}}\mathcal{L}_{\text{kp}}
+
\lambda_{\text{pos}}\mathcal{L}_{\text{pos}}
+
\lambda_{\text{quat}}\mathcal{L}_{\text{quat}}.
\]

This objective ensures that the decoded skeleton faithfully reconstructs the geometric structure of the motion while producing physically valid joint rotations.

\begin{algorithm}[t]
\caption{Point Cloud Tokenizer (VQ-VAE)}
\begin{algorithmic}[1]
\REQUIRE Point cloud sequence $\mathbf{X}\in\mathbb{R}^{F\times N\times3}$
\STATE Encode each frame $\mathbf{Z}=\text{Encoder}(\mathbf{X})\in\mathbb{R}^{F\times d}$
\STATE Quantize latents via codebook
$\mathbf{z}_f^q=\mathbf{e}_{k^*},\; k^*=\arg\min_k\|\mathbf{z}_f-\mathbf{e}_k\|^2$
\STATE Obtain token sequence $\mathbf{Z}^q=[\mathbf{z}_1^q,\dots,\mathbf{z}_F^q]$
\STATE Reconstruct point cloud
$\hat{\mathbf{X}}=\text{Decoder}(\mathbf{Z}^q)$
\RETURN quantized tokens $\mathbf{Z}^q$
\end{algorithmic}
\end{algorithm}

\subsection{Point Cloud Tokenizer}

\subsubsection{Tokenizer Encoder}

Given a point cloud sequence \(\mathbf{X} \in \mathbb{R}^{F \times N \times 3}\), the VQ Encoder first extracts per-frame latent features by encoding the spatial dimension of each frame:
\[
    \mathbf{Z} = \mathrm{Encoder}(\mathbf{X}) \in \mathbb{R}^{F \times d},
\]
where each frame’s \(N \times 3\) points are encoded into a \(d\)-dimensional latent vector, preserving the temporal dimension \(F\).

Next, vector quantization is applied along the temporal dimension to discretize \(\mathbf{Z}\). For each latent vector \(\mathbf{z}_f \in \mathbb{R}^d\), its nearest codebook embedding \(\mathbf{e}_{k^*}\) is selected from the learnable codebook \(\mathcal{C} = \{\mathbf{e}_1, \dots, \mathbf{e}_K\} \subset \mathbb{R}^d\):
\[
    k^* = \arg\min_{k} \| \mathbf{z}_f - \mathbf{e}_k \|_2^2, \quad \mathbf{z}_f^q = \mathbf{e}_{k^*}.
\]
This yields the quantized latent sequence:
\[
    \mathbf{Z}^q = [\mathbf{z}_1^q, \mathbf{z}_2^q, \dots, \mathbf{z}_K^q] \in \mathbb{R}^{K \times d}.
\]
Here, \(K\) is the number of codebook entries, and each \(\mathbf{z}_k^q\) is the embedding of a learned discrete token.

\subsubsection{Tokenizer Decoder}

The quantized latent tokens \(\mathbf{Z}^q\) are decoded back to reconstruct the original point cloud sequence by the Tokenizer Decoder:
\[
    \hat{\mathbf{X}} = \mathrm{Decoder}(\mathbf{Z}^q) \in \mathbb{R}^{F \times N \times 3}.
\]

\subsubsection{Learning objective.}  The training objective combines three loss terms:
\[
    \mathcal{L} = \mathcal{L}_{\text{vertice}} + \lambda_{\text{commit}} \mathcal{L}_{\text{commit}} + \lambda_{\text{identity}} \mathcal{L}_{\text{identity}},
\]
where \(\mathcal{L}_{\text{vertice}}\) measures reconstruction error between \(\mathbf{X}\) and \(\hat{\mathbf{X}}\), e.g., via Chamfer Distance. The commitment loss \(\mathcal{L}_{\text{commit}}\) encourages the encoder outputs \(\mathbf{Z}\) to commit to the quantized embeddings \(\mathbf{Z}^q\), while \(\mathcal{L}_{\text{identity}}\) stabilizes codebook learning. The hyperparameters \(\lambda_{\text{commit}}\) and \(\lambda_{\text{identity}}\) balance these losses.

\begin{algorithm}[t]
\caption{Mask Transformer for Motion Token Modeling}
\begin{algorithmic}[1]
\REQUIRE Quantized token sequence $\mathbf{Z}^q=[\mathbf{z}_1^q,\dots,\mathbf{z}_K^q]$
\STATE Add positional encoding $\widetilde{\mathbf{z}}_k=\mathbf{z}_k^q+\mathrm{PE}(k)$
\STATE Randomly mask token subset $\mathcal{M}$
\STATE Contextualize tokens
$\mathbf{H}=\text{Transformer}(\widetilde{\mathbf{Z}})$
\FOR{$k\in\mathcal{M}$}
\STATE Predict code index $\mathbf{L}_k=\mathbf{W}_{out}\mathbf{h}_k+\mathbf{b}_{out}$
\STATE Compute masked token loss $\mathcal{L}_{CE}$
\ENDFOR
\RETURN predicted token sequence
\end{algorithmic}
\end{algorithm}

\begin{table*}[t]
\caption{\textbf{Evaluation on UniML3D.} The right arrow $\rightarrow$ means that the closer to the real motion, the better. \textbf{Bold} indicates best results.}
\label{tab:uniml3d}
\resizebox{0.8\linewidth}{!}{%
\begin{tabular}{@{}lccccccc@{}}
\toprule
\multirow{2}{*}{Method} & \multicolumn{3}{c}{R Precision $\uparrow$}                             & \multicolumn{1}{c}{\multirow{2}{*}{FID$\downarrow$}} & \multirow{2}{*}{MM Dist$\downarrow$} & \multirow{2}{*}{Diversity$\rightarrow$} & \multirow{2}{*}{MModality$\uparrow$} \\
\cmidrule(lr){2-4}
                        & Top\,1              & Top\,2              & Top\,3              &                              &                                      &                                        &                                      \\
\midrule
\multicolumn{8}{c}{\textbf{Whole Dataset}}\\
\midrule
Real                    & $0.565^{\pm0.004}$   & $0.741^{\pm0.005}$   & $0.846^{\pm0.004}$   & $0.011^{\pm0.001}$& $2.458^{\pm0.006}$                    & $9.416^{\pm0.065}$& $2.812^{\pm0.046}$                    \\
\midrule
T2M-GPT~\cite{zhang2023generating}                 & $0.500^{\pm0.004}$ & $0.700^{\pm0.006}$ & $0.785^{\pm0.004}$ & $0.095^{\pm0.014}$          & $2.780^{\pm0.010}$                  & $9.640^{\pm0.075}$                      & $2.200^{\pm0.060}$                  \\
MotionGPT~\cite{jiang2023motiongpt}                  & $0.530^{\pm0.002}$ & $0.730^{\pm0.003}$ & $0.835^{\pm0.005}$ & $0.070^{\pm0.005}$          & $2.750^{\pm0.006}$                  & $9.590^{\pm0.070}$                      & $2.450^{\pm0.085}$                  \\
MDM~\cite{tevet2022human}                      & $0.540^{\pm0.004}$ & $0.742^{\pm0.004}$ & $0.838^{\pm0.002}$ & $0.058^{\pm0.003}$          & $2.740^{\pm0.007}$                  & $\textbf{9.587}^{\pm0.044}$             & $2.600^{\pm0.040}$                  \\
MotionDiffuse~\cite{zhang2024motiondiffuse}             & $0.545^{\pm0.003}$ & $0.745^{\pm0.004}$ & $0.840^{\pm0.006}$ & $0.055^{\pm0.003}$          & $2.730^{\pm0.005}$                  & $9.592^{\pm0.078}$                      & $2.650^{\pm0.065}$                  \\
OmniMotionGPT~\cite{yang2024omnimotiongpt}                & $0.505^{\pm0.006}$ & $0.712^{\pm0.003}$ & $0.802^{\pm0.006}$ & $0.090^{\pm0.009}$          & $2.785^{\pm0.002}$                  & $9.635^{\pm0.058}$                      & $2.225^{\pm0.068}$                  \\
Motion Avatar~\cite{zhang2024motionavatar}             & $0.552^{\pm0.005}$ & $0.750^{\pm0.002}$ & $0.845^{\pm0.004}$ & $0.047^{\pm0.003}$          & $2.725^{\pm0.012}$                  & $9.600^{\pm0.050}$                      & $2.705^{\pm0.030}$                  \\

\midrule
\textbf{UniMo (Ours)}   & $\textbf{0.564}^{\pm0.005}$ & $\textbf{0.755}^{\pm0.006}$ & $\textbf{0.848}^{\pm0.005}$ & $\textbf{0.040}^{\pm0.002}$ & $\textbf{2.710}^{\pm0.006}$ & $9.620^{\pm0.056}$ & $\textbf{2.800}^{\pm0.046}$ \\
\midrule
\multicolumn{8}{c}{\textbf{Animal Only}}\\
\midrule
Real                    & $0.563^{\pm0.006}$& $0.739^{\pm0.007}$& $0.841^{\pm0.005}$& $0.013^{\pm0.002}$& $2.193^{\pm0.005}$& $9.503^{\pm0.067}$& $2.815^{\pm0.039}$\\
\midrule
T2M-GPT~\cite{zhang2023generating}                 & $0.495^{\pm0.005}$ & $0.692^{\pm0.006}$ & $0.775^{\pm0.005}$ & $0.100^{\pm0.013}$          & $2.805^{\pm0.010}$                  & $9.630^{\pm0.072}$                      & $2.190^{\pm0.061}$                  \\
MotionGPT~\cite{jiang2023motiongpt}                  & $0.525^{\pm0.003}$ & $0.725^{\pm0.005}$ & $0.830^{\pm0.005}$ & $0.068^{\pm0.007}$          & $2.770^{\pm0.008}$                  & $9.585^{\pm0.070}$                      & $2.455^{\pm0.087}$                  \\
MDM~\cite{tevet2022human}                      & $0.535^{\pm0.005}$ & $0.740^{\pm0.005}$ & $0.838^{\pm0.002}$ & $0.058^{\pm0.003}$          & $2.765^{\pm0.008}$                  & $\textbf{9.580}^{\pm0.044}$                      & $2.590^{\pm0.039}$                  \\
MotionDiffuse~\cite{zhang2024motiondiffuse}             & $0.540^{\pm0.004}$ & $0.743^{\pm0.004}$ & $0.842^{\pm0.005}$ & $0.056^{\pm0.002}$          & $2.760^{\pm0.006}$                  & $9.585^{\pm0.077}$                      & $2.650^{\pm0.064}$                  \\
OmniMotionGPT~\cite{yang2024omnimotiongpt}                & $0.500^{\pm0.006}$ & $0.710^{\pm0.004}$ & $0.800^{\pm0.006}$ & $0.090^{\pm0.008}$          & $2.790^{\pm0.002}$                  & $9.630^{\pm0.057}$                      & $2.220^{\pm0.067}$                  \\
Motion Avatar~\cite{zhang2024motionavatar}             & $0.550^{\pm0.006}$ & $0.748^{\pm0.003}$ & $0.843^{\pm0.003}$ & $0.049^{\pm0.002}$          & $2.755^{\pm0.013}$                  & $9.600^{\pm0.047}$                      & $2.700^{\pm0.029}$                  \\

\midrule
\textbf{UniMo (Ours)}   & $\textbf{0.560}^{\pm0.006}$ & $\textbf{0.753}^{\pm0.005}$ & $\textbf{0.847}^{\pm0.005}$ & $\textbf{0.042}^{\pm0.002}$ & $\textbf{2.740}^{\pm0.007}$ & $9.615^{\pm0.056}$ & $\textbf{2.800}^{\pm0.046}$ \\
\bottomrule
\end{tabular}%
}
\end{table*}

\subsection{Mask Transformer}
To model temporal dependencies in the quantized latent sequence and enable robust motion generation, we use a Mask Transformer. Given the quantized latent sequence \(\mathbf{Z}^q = [\mathbf{z}_1^q, \dots, \mathbf{z}_K^q] \in \mathbb{R}^{K \times d}\), where each \(\mathbf{z}_k^q\) is a codebook embedding, the goal is to learn contextualized representations that encode the dynamics across tokens.

We first add sinusoidal positional encoding to each token to incorporate order information:
\[
    \widetilde{\mathbf{z}}_k = \mathbf{z}_k^q + \mathrm{PE}(k), \quad k = 1, \dots, K.
\]

Let \(\widetilde{\mathbf{Z}} = [\widetilde{\mathbf{z}}_1, \dots, \widetilde{\mathbf{z}}_K] \in \mathbb{R}^{K \times d}\) be the sequence after encoding. We feed it into a multi-layer Transformer encoder:
\[
    \mathbf{H} = \mathrm{Transformer}(\widetilde{\mathbf{Z}}) \in \mathbb{R}^{K \times d},
\]
where \(\mathbf{H} = [\mathbf{h}_1, \dots, \mathbf{h}_K]\) are the contextualized token representations.

Each \(\mathbf{h}_k\) is projected to a logit vector over the codebook indices:
\[
    \mathbf{L}_k = \mathbf{W}_{\mathrm{out}} \mathbf{h}_k + \mathbf{b}_{\mathrm{out}} \in \mathbb{R}^{K}, \quad k = 1, \dots, K,
\]
which corresponds to the predicted distribution over token indices at position \(k\).

During training, a random subset of token positions is masked using a BERT-style masking scheme. The objective is to reconstruct the original codebook indices from context:
\[
    \mathcal{L}_{\mathrm{CE}} = -\sum_{k \in \mathcal{M}} \log P(m_k \mid \mathbf{L}_k),
\]
where \(\mathcal{M}\) is the set of masked positions and \(m_k \in \{1, \dots, K\}\) is the ground-truth codebook index at position \(k\).

This Mask Transformer enables sequence modeling over discrete motion tokens, capturing long-range dependencies and supporting diverse motion generation.

\section{Experiments}

\subsection{Public Benchmarks and Evaluation Metrics}

\subsubsection{Benchmarks}

To ensure a fair comparison, we evaluate our method on both our UniML3D dataset and public benchmarks. For human motion generation, we use standard datasets including HumanML3D~\cite{guo2022generating} and KIT-ML~\cite{plappert2016kit}. For animal motion generation, we use AnimalML3D~\cite{yang2024omnimotiongpt}.

\begin{table*}[t]
\caption{\textbf{Evaluation on AnimalML3D \cite{yang2024omnimotiongpt}, HumanML3D \cite{guo2022generating}, and KIT-ML \cite{plappert2016kit}.} The right arrow $\rightarrow$ means that the closer to the real motion, the better. \textbf{Bold} and \underline{underline} indicate best and second best results.}
\label{tab:animl3d}
\resizebox{0.75\linewidth}{!}{%
\begin{tabular}{@{}lccccccc@{}}
\toprule
\multirow{2}{*}{Method} & \multicolumn{3}{c}{R Precision $\uparrow$}                                                                                                                & \multicolumn{1}{c}{\multirow{2}{*}{FID$\downarrow$}} & \multirow{2}{*}{MM Dist$\downarrow$}              & \multirow{2}{*}{Diversity$\rightarrow$}           & \multirow{2}{*}{MModality$\uparrow$}              \\ \cmidrule(lr){2-4}
              & \multicolumn{1}{c}{Top 1} & \multicolumn{1}{c}{Top 2} & \multicolumn{1}{c}{Top 3} & \multicolumn{1}{c}{}                     &                          &                            &                            \\
\midrule
\multicolumn{8}{c}{\textbf{AnimalML3D \cite{yang2024omnimotiongpt}}}\\
\midrule
Real &
\et{ 0.558}{ .049}      & %
\et{ 0.734}{ .040}      & %
\et{ 0.839}{ .032}      & %
\et{ 0.105}{ .005}      & %
\et{ 0.357}{ .006}      & %
\et{22.795}{1.843}      & %
-                       \\
  \midrule
        T2M-GPT~\cite{zhang2023generating} &
        \et{ 0.080}{ .024}          & %
        \et{ 0.168}{ .023}          & %
        \et{ 0.248}{ .042}          & %
        \et{ 1.084}{ .042}          & %
        \et{ 0.636}{ .013}          & %
        \et{33.403}{1.902}        & %
        \et{20.078}{1.096}         \\

        MotionGPT~\cite{jiang2023motiongpt} &
        \et{ 0.142}{ .016}                  & %
        \et{ 0.233}{ .032}                  & %
        \et{ 0.307}{ .042}                  & %
        \et{ 0.748}{ .050}                  & %
        \et{ 0.558}{ .010}                  & %
        \et{29.265}{2.453}                  & %
        \et{10.311}{1.537}                  \\

        MDM~\cite{tevet2022human}   & 
        \et{ 0.379}{ .051}          & %
        \et{ 0.554}{ .058}          & %
        \et{ 0.646}{ .048}          & %
        \et{ 0.505}{ .038}          & %
        \et{ 0.487}{ .008}          & %
        \et{\underline{27.826}}{1.643}          & %
        \et{13.593}{1.038}          \\

        MotionDiffuse~\cite{zhang2024motiondiffuse} & 
        \et{ 0.505}{ .037}                        & %
        \et{ 0.695}{ .045}                        & %
        \et{ 0.805}{ .041}                        & %
        \et{ 0.401}{ .024}                        & %
        \et{ 0.421}{ .007}                        & %
        \et{\textbf{25.194}}{1.510}                         & %
        \et{ 7.081}{0.357}                          \\
OmniMotionGPT \cite{yang2024omnimotiongpt}       &
        \et{ 0.539}{ .064} & %
        \et{ 0.721}{ .063} & %
        \et{ 0.830}{ .043} & %
        \et{ 0.223}{ .036} & %
        \et{ 0.348}{ .007} & %
        \et{37.487}{1.575} & %
        \et{17.487}{0.792}\\
Motion Avatar \cite{zhang2024motionavatar}       &
        \et{ \underline{0.544}}{ .053} & %
        \et{ \underline{0.726}}{ .062} & %
        \et{ \underline{0.833}}{ .031} & %
        \et{ \underline{0.159}}{ .048} & %
        \et{ \underline{0.320}}{ .009} & %
        \et{34.593}{1.326} & %
        \et{\underline{18.463}}{0.849}\\
\midrule
\textbf{UniMo (Ours)} &
$\textbf{0.547}^{\pm.003}$ &
$\textbf{0.731}^{\pm.001}$ &
$\textbf{0.836}^{\pm.004}$ &
$\textbf{0.103}^{\pm.006}$ &
$\textbf{0.317}^{\pm.003}$ &
${31.364}^{\pm.033}$ &
$\textbf{18.831}^{\pm.041}$ \\
\midrule
\multicolumn{8}{c}{\textbf{HumanML3D \cite{guo2022generating}}}\\
\midrule
Real &
 $0.511^{\pm.003}$ &
 $0.703^{\pm.003}$ &
 $0.797^{\pm.002}$ &
 $0.002^{\pm.000}$ &
 $2.974^{\pm.008}$ &
 $9.503^{\pm.065}$ &
  -
  \\ \midrule
ReMoDiffuse \cite{zhang2023remodiffuse} &
    ${0.510}^{\pm.005}$ &
    ${0.698}^{\pm.006}$ &
    ${0.795}^{\pm.004}$ &
    ${0.103}^{\pm.004}$ &
    ${2.974}^{\pm.016}$ &
    ${9.018}^{\pm.075}$ &
    $1.795^{\pm.043}$ \\ 
MMM \cite{pinyoanuntapong2024mmm} &
    ${0.504}^{\pm.003}$ &
    ${0.696}^{\pm.003}$ &
    ${0.794}^{\pm.002}$ &
    ${0.080}^{\pm.003}$ &
    ${2.998}^{\pm.007}$ &
    $\underline{9.411}^{\pm.058}$ &
    $1.164^{\pm.041}$ \\ 
DiverseMotion \cite{lou2023diversemotion} &
    ${0.515}^{\pm.003}$ &
    ${0.706}^{\pm.002}$ &
    ${0.802}^{\pm.002}$ &
    ${0.072}^{\pm.004}$ &
    ${2.941}^{\pm.007}$ &
    ${9.683}^{\pm.102}$ &
    $\underline{1.869}^{\pm.089}$ \\ 
BAD \cite{hosseyni2024bad} &
    ${0.517}^{\pm.002}$ &
    ${0.713}^{\pm.003}$ &
    ${0.808}^{\pm.003}$ &
    ${0.065}^{\pm.003}$ &
    ${2.901}^{\pm.008}$ &
    ${9.694}^{\pm.068}$ &
    $1.194^{\pm.044}$ \\
BAMM \cite{pinyoanuntapong2024bamm} &
    ${0.525}^{\pm.002}$ &
    ${0.720}^{\pm.003}$ &
    ${0.814}^{\pm.003}$ &
    ${0.055}^{\pm.002}$ &
    ${2.919}^{\pm.008}$ &
    ${9.717}^{\pm.089}$ &
    $1.687^{\pm.051}$ \\
MoMask \cite{guo2024momask} &
    ${0.521}^{\pm.002}$ &
    ${0.713}^{\pm.002}$ &
    ${0.807}^{\pm.002}$ &
    ${0.045}^{\pm.002}$ &
    ${2.958}^{\pm.008}$ &
    - &
    $1.241^{\pm.040}$ \\
MoGenTS \cite{yuan2024mogents} &
    $\underline{0.529}^{\pm.003}$ &
    $\underline{0.719}^{\pm.002}$ &
    $\underline{0.812}^{\pm.002}$ &
    $\underline{0.033}^{\pm.001}$ &
    $\underline{2.867}^{\pm.006}$ &
    $\textbf{9.570}^{\pm.077}$ &
    - \\
 \midrule
\textbf{UniMo (Ours)} &
$\textbf{0.533}^{\pm.003}$ &
$\textbf{0.720}^{\pm.006}$ &
$\textbf{0.816}^{\pm.002}$ &
$\textbf{0.031}^{\pm.001}$ &
$\textbf{2.859}^{\pm.006}$ &
${9.625}^{\pm.035}$ &
$\textbf{1.901}^{\pm.077}$ \\
\midrule
\multicolumn{8}{c}{\textbf{KIT-ML \cite{plappert2016kit}}}\\
\midrule
Real &
 $0.424^{\pm.005}$ &
 $0.649^{\pm.006}$ &
 $0.779^{\pm.006}$ &
 $0.031^{\pm.004}$ &
 $2.788^{\pm.012}$ &
 $11.08^{\pm.097}$ &
  -
  \\ \midrule
ReMoDiffuse \cite{zhang2023remodiffuse} &
    ${0.427}^{\pm.014}$ &
    ${0.641}^{\pm.004}$ &
    ${0.765}^{\pm.055}$ &
    ${0.155}^{\pm.006}$ &
    ${2.814}^{\pm.012}$ &
    ${10.80}^{\pm.105}$ &
    $1.239^{\pm.028}$ \\ 

MMM \cite{pinyoanuntapong2024mmm} &
    ${0.404}^{\pm.005}$ &
    ${0.621}^{\pm.005}$ &
    ${0.744}^{\pm.004}$ &
    ${0.316}^{\pm.028}$ &
    ${2.977}^{\pm.019}$ &
    ${10.91}^{\pm.101}$ &
    $1.232^{\pm.039}$ \\ 

DiverseMotion \cite{lou2023diversemotion} &
    ${0.416}^{\pm.005}$ &
    ${0.637}^{\pm.008}$ &
    ${0.760}^{\pm.011}$ &
    ${0.468}^{\pm.098}$ &
    ${2.892}^{\pm.041}$ &
    ${10.87}^{\pm.101}$ &
    $\underline{2.062}^{\pm.079}$ \\ 

BAD \cite{hosseyni2024bad} &
    ${0.417}^{\pm.006}$ &
    ${0.631}^{\pm.006}$ &
    ${0.750}^{\pm.006}$ &
    ${0.221}^{\pm.012}$ &
    ${2.941}^{\pm.025}$ &
    $\underline{11.00}^{\pm.100}$ &
    $1.170^{\pm.047}$ \\

BAMM \cite{pinyoanuntapong2024bamm} &
    ${0.438}^{\pm.009}$ &
    ${0.661}^{\pm.009}$ &
    ${0.788}^{\pm.005}$ &
    ${0.183}^{\pm.013}$ &
    ${2.723}^{\pm.026}$ &
    $\textbf{11.01}^{\pm.094}$ &
    $1.609^{\pm.065}$ \\

MoMask \cite{guo2024momask} &
    ${0.433}^{\pm.007}$ &
    ${0.656}^{\pm.005}$ &
    ${0.781}^{\pm.005}$ &
    ${0.204}^{\pm.011}$ &
    ${2.779}^{\pm.022}$ &
    - &
    $1.131^{\pm.043}$ \\

MoGenTS \cite{yuan2024mogents} &
    $\textbf{0.445}^{\pm.006}$ &
    $\textbf{0.671}^{\pm.006}$ &
    $\textbf{0.797}^{\pm.005}$ &
    $\underline{0.143}^{\pm.004}$ &
    $\underline{2.711}^{\pm.024}$ &
    $10.92^{\pm.090}$ &
    - \\
 \midrule
\textbf{UniMo (Ours)} &
$\underline{0.441}^{\pm.003}$ &
$\underline{0.668}^{\pm.004}$ &
$\underline{0.795}^{\pm.006}$ &
$\textbf{0.139}^{\pm.001}$ &
$\textbf{2.709}^{\pm.003}$ &
${10.77}^{\pm.048}$ &
$\textbf{2.081}^{\pm.077}$ \\
\bottomrule
\end{tabular}%
   }
\end{table*}

\subsubsection{Evaluation metrics.}

We adopt standard text-to-motion metrics \cite{guo2022generating,yang2024omnimotiongpt} to assess different aspects of our experiments. We use FID and R-Precision to evaluate the realism and accuracy of generated motions. MM-Dist measures motion-text alignment, while a diversity metric quantifies variation in motion features. Additionally, the MModality metric assesses the diversity of motions generated from the same text description.

\subsection{Implementation Details}

To ensure fair comparison, both the baselines and our model are trained from scratch on each benchmark.
We employ three training stages: the skeleton decoder comprises a 6-layer Transformer block with 4 attention heads, a learning rate of $1 \times 10^{-4}$; the point-cloud autoencoder uses three 1D convolutional layers for spatial processing followed by a VQ-VAE that compresses the temporal dimension with code size = 512, and codebook dimension = 256, with a learning rate of $1 \times 10^{-4}$; the Mask Transformer has a depth of 6 layers, 8 attention heads, dropout rate = 0.2, latent dimension = 384, and a learning rate of $2 \times 10^{-1}$. We employ a frozen text encoder from CLIP ViT-B/32. A batch size of 256 and a maximum of 5K epochs are used for each stage. All experiments are conducted on a single NVIDIA A100 40G GPU.

\subsection{Comparative Study}

Based on the results in Table~\ref{tab:uniml3d} and Table~\ref{tab:animl3d}, our method, \textbf{UniMo}, achieves state-of-the-art performance across all datasets, including UniML3D, AnimalML3D, HumanML3D, and KIT-ML. On UniML3D, UniMo consistently outperforms all baselines in R-Precision, FID, MM-Dist, Diversity, and MModality, both on the full dataset and the animal-only subset. Similarly, on AnimalML3D, UniMo matches or surpasses all prior methods in motion-text alignment and diversity. On HumanML3D and KIT-ML, UniMo remains competitive and achieves the best or second-best scores across most metrics, validating its effectiveness for both human and animal motion generation within a unified framework.

\begin{table*}[t]
\caption{\textbf{Number of point cloud.} The right arrow $\rightarrow$ means that the closer to the real motion, the better. \textbf{Bold} indicate best.}
\label{tab:point}
\resizebox{0.8\linewidth}{!}{%
\begin{tabular}{@{}lccccccc@{}}
\toprule
\multirow{2}{*}{Number of point cloud} & \multicolumn{3}{c}{R Precision $\uparrow$}                             & \multicolumn{1}{c}{\multirow{2}{*}{FID$\downarrow$}} & \multirow{2}{*}{MM Dist$\downarrow$} & \multirow{2}{*}{Diversity$\rightarrow$} & \multirow{2}{*}{MModality$\uparrow$} \\
\cmidrule(lr){2-4}
                        & Top\,1              & Top\,2              & Top\,3              &                              &                                      &                                        &                                      \\
\midrule                        
Real                   & $0.565^{\pm0.004}$   & $0.741^{\pm0.005}$   & $0.846^{\pm0.004}$   & $0.011^{\pm0.001}$ & $2.458^{\pm0.006}$                    & $9.416^{\pm0.065}$ & $2.812^{\pm0.046}$                    \\
\midrule
64                 & $0.495^{\pm0.008}$ & $0.700^{\pm0.010}$ & $0.780^{\pm0.007}$ & $0.100^{\pm0.015}$          & $2.780^{\pm0.010}$                  & $9.600^{\pm0.080}$                      & $2.200^{\pm0.070}$                  \\
128                  & $0.525^{\pm0.006}$ & $0.730^{\pm0.007}$ & $0.835^{\pm0.006}$ & $0.070^{\pm0.010}$          & $2.760^{\pm0.008}$                  & $\textbf{9.600}^{\pm0.075}$                      & $2.500^{\pm0.065}$                  \\
256                      & $\textbf{0.564}^{\pm0.005}$ & $\textbf{0.755}^{\pm0.006}$ & $0.848^{\pm0.005}$ & $\textbf{0.040}^{\pm0.002}$ & ${2.710}^{\pm0.006}$                  & $9.620^{\pm0.056}$             & $\textbf{2.800}^{\pm0.046}$                  \\
384             & $0.560^{\pm0.004}$ & $0.752^{\pm0.005}$ & $0.846^{\pm0.004}$ & $0.042^{\pm0.002}$          & $2.715^{\pm0.005}$                  & $9.615^{\pm0.060}$                      & $2.790^{\pm0.045}$                  \\
512                & $0.560^{\pm0.005}$ & $0.754^{\pm0.006}$ & $\textbf{0.850}^{\pm0.004}$ & $0.041^{\pm0.002}$          & $\textbf{2.705}^{\pm0.005}$                  & $9.610^{\pm0.055}$                      & $2.790^{\pm0.045}$                  \\
\bottomrule
\end{tabular}%
}
\end{table*}

\subsection{Ablation Study}

\subsubsection{Number of point cloud.}

To evaluate the impact of point cloud resolution on motion generation, we conduct an ablation study by varying the number of sampled points $N$ from 64 to 512. As shown in Table~\ref{tab:point}, performance improves steadily with more points, particularly in terms of R Precision, FID, and MModality. The best overall performance is achieved at $N=256$, where our model attains an R Precision Top-3 score of \textbf{0.848}, FID of \textbf{0.040}, and MModality of \textbf{2.800}. Increasing $N$ beyond this provides marginal gains or saturation. These results suggest that a moderate number of well-sampled points are sufficient to capture both motion quality and multimodal alignment, while also maintaining computational efficiency.

\subsubsection{Model configuration.}
To validate the effectiveness of our architectural design, we conduct ablation studies on key model components, including the number of Transformer layers, codebook size, and latent dimension. Reducing the Mask Transformer depth from 6 to 3 layers leads to a drop in R Precision@3 from \textbf{0.848} to 0.823 and an increase in FID from \textbf{0.040} to 0.056. Similarly, using a smaller codebook size of 256 instead of 512 reduces MModality from \textbf{2.800} to 2.645. Lowering the latent dimension from 384 to 256 results in degraded performance across all metrics, with FID rising to 0.062 and R Precision@3 falling to 0.819. These results confirm that our final configuration—6-layer skeleton decoder, VQ-VAE with code size 512 and dimension 256, and 6-layer Mask Transformer with 384-dimensional latents—strikes the best balance between performance and efficiency. All models are trained from scratch under consistent settings on each benchmark using a batch size of 256 and up to 5000 epochs on a single NVIDIA A100 40G GPU.

\subsection{Efficiency}

\begin{figure}[t]
  \centering
  \begin{subfigure}[b]{0.49\linewidth}
    \includegraphics[width=\linewidth]{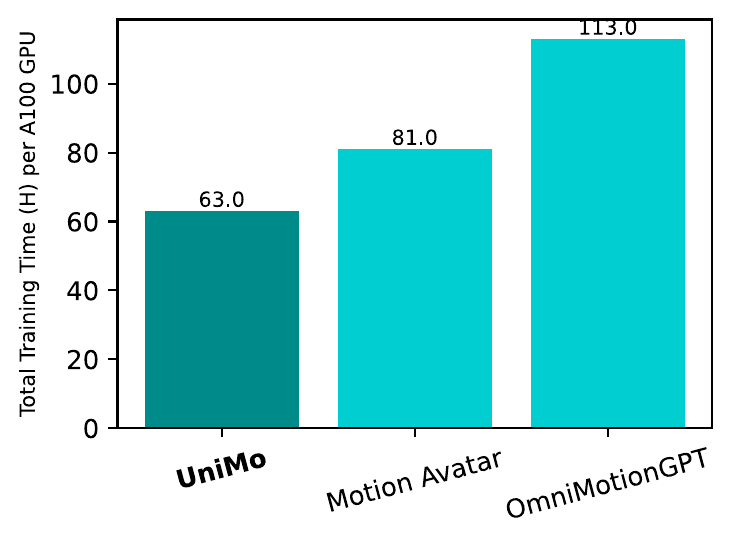}
    \caption{Training Time $\downarrow$}
  \end{subfigure}\hfill
  \begin{subfigure}[b]{0.49\linewidth}
    \includegraphics[width=\linewidth]{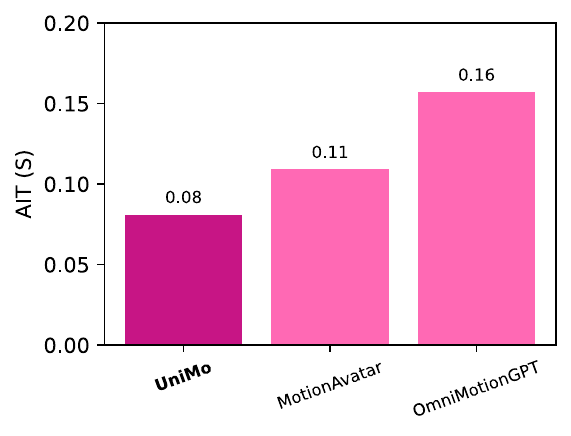}
    \caption{AIT(s) $\downarrow$}
  \end{subfigure}
  \caption{\textbf{Efficiency comparison.} The figure demonstrates that UniMo achieves the lowest inference time and training time, while maintaining superior performance compared to other methods.}
  \label{fig:efficiency}
\end{figure}

Figure~\ref{fig:efficiency} highlights the efficiency advantages of \textbf{UniMo} over prior methods. Specifically, UniMo requires only \textbf{63 hours} of training on a single A100 GPU, significantly faster than Motion Avatar (\textbf{81 hours}) and OmniMotionGPT (\textbf{113 hours}). Moreover, UniMo achieves the lowest autoregressive inference time (AIT), with a latency of just \textbf{0.08 seconds} per frame, compared to \textbf{0.11 seconds} for Motion Avatar and \textbf{0.16 seconds} for OmniMotionGPT. These results demonstrate that UniMo not only improves performance but also greatly enhances training and inference efficiency, making it more practical for real-world deployment.

\section{Limitations and Future Work}

Although UniML3D is significantly larger than existing animal motion datasets, it is still limited to curated motion sequences and does not incorporate motion data extracted from real-world videos. This constrains its scalability and diversity, particularly for capturing naturalistic and in-the-wild animal behaviors. In future work, we aim to explore large-scale dataset construction by leveraging video sources and applying motion reconstruction techniques to extract 3D motion data automatically. This direction could further enhance the realism, variety, and generalization capacity of unified motion generation models.

\section{Conclusion}

In this work, we present \textbf{UniMo}, a unified point cloud-based motion generation framework capable of handling diverse skeletal topologies across human and animal categories. By converting parametric skeletons into unparametric point cloud representations and introducing a dynamic sampling strategy, UniMo enables flexible and scalable motion synthesis across species. Furthermore, we construct \textbf{UniML3D}, the largest unified motion-language dataset to date, significantly advancing data availability for both human and animal motion generation. Extensive experiments on UniML3D and standard benchmarks demonstrate the superiority of our method over existing approaches, highlighting the potential of point-based modeling for generalizable and high-quality motion generation.

\clearpage
\bibliographystyle{ACM-Reference-Format}
\bibliography{sample-sigconf}

\clearpage
\appendix

\section{User Study and Quantative Evaluation}

We conduct a comprehensive user study to evaluate the real-world applicability and perceptual quality of motion sequences generated by our method in comparison with Motion Avatar~\cite{zhang2024motionavatar} and OmniMotionGPT~\cite{yang2024omnimotiongpt}. A total of 50 participants completed a Google Forms survey designed to assess the quality, diversity, and alignment of the generated motions.

\begin{figure}[b]
  \centering
  \includegraphics[width=0.9\linewidth]{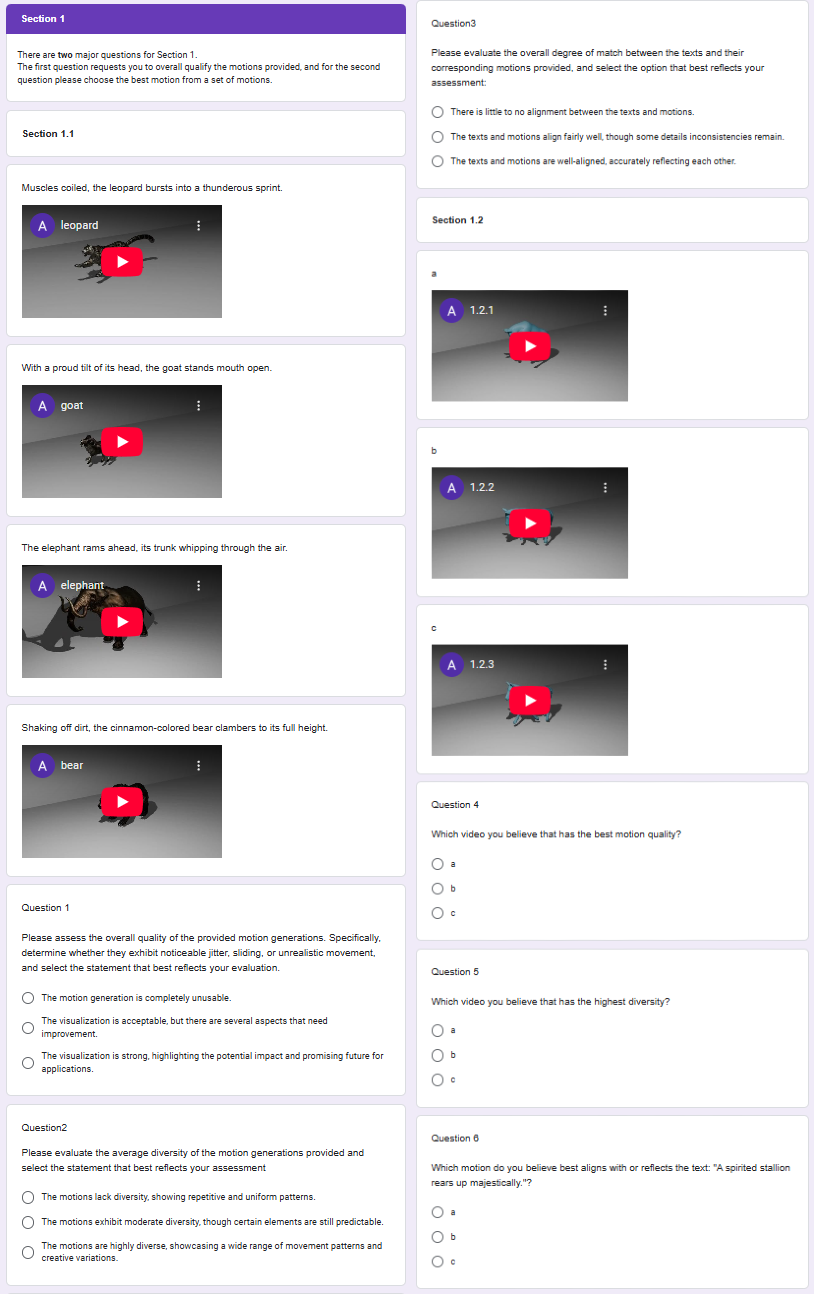}
  \caption{\textbf{User study Google Forms.} The User Interface (UI) used in our user study.}
  \label{fig:user_study_ui}
\end{figure}

As illustrated in Figure~\ref{fig:user_study_ui}, the user interface displays 3–4 motion clips (Videos 1–3/4) generated by the same model, followed by a comparison set (Videos A–C) from different models. Participants rated each animation on a 3-point Likert scale (1 = low, 3 = high) based on motion accuracy and overall visual experience. In the comparison section, users selected the motion sequence they found most realistic and engaging.

This study aims to evaluate not only the fidelity of the generated motions to real-world animal movement but also the overall effectiveness of each model in producing visually compelling results.

\paragraph{Results.}
\begin{itemize}
    \item Our method achieved a motion quality rating of \textbf{2.90}, with \textbf{92\%} of participants agreeing that it produces high-quality motion with minimal jitter, sliding, or unrealistic artifacts.
    \item For motion diversity, we received a rating of \textbf{2.88}, with \textbf{88\%} of participants indicating that our method generates complex and varied motion sequences.
    \item In terms of text-motion alignment, our model scored \textbf{2.96}, and \textbf{96\%} of users reported that the generated motions were well-aligned with the given text descriptions.
    \item Notably, \textbf{94\%} of participants preferred our method over the baselines in the pairwise comparison.
\end{itemize}

\end{document}